\documentclass{article}

\usepackage{microtype}
\usepackage{etoolbox}
\usepackage{graphicx}
\usepackage{xcolor}
\usepackage{booktabs}
\usepackage{capt-of}
\usepackage{needspace}
\usepackage{amsmath,amssymb}
\usepackage{array}
\usepackage{tikz}
\usetikzlibrary{arrows.meta,positioning,matrix,fit,calc,decorations.pathreplacing}
\usepackage{hyperref}

\usepackage{mlsys2025}
\makeatletter
\gdef\isaccepted{1}
\gdef\Notice@String{}
\makeatother
\hypersetup{hypertexnames=false,pdfsubject={arXiv preprint}}
\AtBeginEnvironment{thebibliography}{\small\raggedright}
\definecolor{navy}{rgb}{0.09,0.25,0.37}
\definecolor{kernelproblue}{HTML}{3568A8}
\definecolor{sassorange}{HTML}{C26D17}
\definecolor{fasmgreen}{HTML}{237A57}
\definecolor{evalgray}{HTML}{58616B}
\newcommand{\revisioncolor}[1]{#1}
\newcommand{\newrevisioncolor}[1]{#1}

\newcommand{\gf}{\mathbb{F}_2}
\newcommand{\sass}{SASS}
\newcommand{\cubin}{\texttt{CUBIN}}
\newcommand{\tool}{F2Asm}
\newsavebox{\blackwellsharedtablebox}

\mlsystitlerunning{Learning Exact NVIDIA SASS Encoders with $\mathbb{F}_2$ Linear Algebra}

\begin{document}
\twocolumn[
\mlsystitle{Learning Exact NVIDIA SASS Encoders with
\texorpdfstring{$\mathbb{F}_2$}{F2} Linear Algebra}

\begin{mlsysauthorlist}
\mlsysauthor{Jiading Gai}{author}
\end{mlsysauthorlist}
\vspace{-6pt}
\begin{center}
\small
\textbf{Code:}
\href{https://github.com/JiadingGai/f2asm}
     {\texttt{https://github.com/JiadingGai/f2asm}}
\end{center}
\mlsysaffiliation{author}{Independent Researcher}
\mlsyscorrespondingauthor{Jiading Gai}{jiading.gai@gmail.com}
\mlsyskeywords{GPU binary rewriting, SASS, exact encoders, binary finite fields, provenance}

\vskip 0.3in

\begin{abstract}
NVIDIA provides a \sass{} disassembler but no public \sass{}
assembler for recent data-center GPUs, limiting controlled machine-code
rewriting.  We present \textbf{\tool{}}, which learns exact 128-bit \sass{} encoders from
paired disassembly and original \cubin{} instruction words.
To our knowledge, \tool{} is the first system to learn \sass{}
instruction encoders as vector-valued affine maps over $\gf$ and the first
open-source NVIDIA \sass{} assembler to support Rubin SM107.  \tool{} uses Gaussian elimination over $\gf$ to
incrementally build a compact basis, detect inconsistencies, and reject inputs
outside the learned span.  \tool{} separates target-specific
control bits, relocation rules, and \cubin{} metadata from its learning
algorithm.  We train encoders for Hopper
SM90/SM90a, Blackwell SM100, and
Rubin SM107 using 3,225
CUBINs from pinned NVIDIA and third-party production libraries, CUDA 13.3
packages, and CUDA 13.4 Developer Preview archives.  In round-trip tests,
\tool{} reassembles each \cubin{}'s disassembled \sass{}, and all compared
executable text sections match the originals exactly.
Joint training with \tool{} yields one shared encoder for five Blackwell SM
targets and another for three Rubin SM targets, providing strong evidence of
a common SASS encoding scheme for instructions shared within each family.
Continual training extends the Rubin encoder to all 17,159 previously
unsupported cuTile and GROMACS queries with 1,504 additional basis rows,
matching the derived lower bound.
\end{abstract}

{\vspace{0.12in}
\begin{minipage}{\textwidth}
  \centering
  \begin{tikzpicture}[
      node distance=4.8mm,
      font=\sffamily\scriptsize,
      flow/.style={-{Latex[length=1.7mm]}, draw=navy,
                   line width=0.55pt},
      feedback/.style={-{Latex[length=1.7mm]}, draw=navy,
                       dashed, line width=0.65pt, rounded corners=3pt},
      stage/.style={rounded corners=2pt, line width=0.75pt, align=center,
                    minimum height=15mm,
                    inner xsep=3pt, inner ysep=2pt, text=black},
      artifact/.style={rounded corners=1.5pt, draw=navy,
                       fill=black!3, align=center,
                       text width=2.15cm, minimum height=10mm,
                       inner xsep=2.5pt, inner ysep=2pt,
                       text=black}]
    \node[stage, draw=kernelproblue, fill=kernelproblue!7,
          text width=3.05cm] (cudaagent) {
      \textbf{1. CUDA coding agent}\\[-1pt]
      \textit{e.g., KernelPro}\\[1pt]
      profile and diagnose\\[-1pt]
      revise CUDA/CuTe\\[-1pt]
      compile, test, and measure};
    \node[artifact, right=of cudaagent] (handoff) {
      \textbf{Validated source-level result}\\[1pt]
      CUDA/CuTe kernel\\[-1pt]
      $+$ compiled CUBIN};
    \node[stage, draw=sassorange, fill=sassorange!7,
          text width=2.85cm, right=of handoff] (sassagent) {
      \textbf{2. Downstream SASS agent}\\[1pt]
      disassemble and profile\\[-1pt]
      analyze machine code\\[-1pt]
      propose constrained edit};
    \node[stage, draw=fasmgreen, fill=fasmgreen!8, line width=1.15pt,
          text width=3.05cm, right=of sassagent] (f2asm) {
      \textbf{3. \tool{} backend}\\[1pt]
      select target/$(f,p)$\\[-1pt]
      check basis support\\[-1pt]
      encode words and rebuild CUBIN};
    \node[stage, draw=evalgray, fill=evalgray!6,
          text width=2.55cm, right=of f2asm] (evaluate) {
      \textbf{4. External validation}\\[1pt]
      semantic correctness\\[-1pt]
      GPU measurement\\[-1pt]
      retain best CUBIN};
    \draw[flow] (cudaagent) -- (handoff);
    \draw[flow] (handoff) -- (sassagent);
    \draw[flow] (sassagent) -- (f2asm);
    \draw[flow] (f2asm) -- (evaluate);
    \draw[feedback] (evaluate.south) -- ++(0,-3.5mm) -|
      node[pos=0.55, below, font=\sffamily\tiny, text=black]
      {validation feedback} (sassagent.south);
  \end{tikzpicture}
  
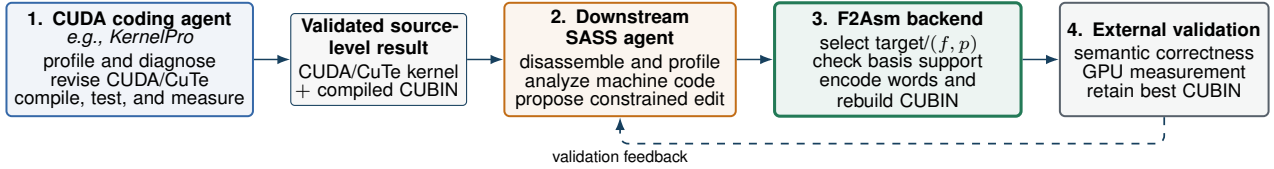
\captionof{figure}{One application of \tool{}.  A CUDA coding agent such as
  KernelPro performs source-level optimization using profiler
  feedback~\cite{DBLP:journals/corr/abs-2606-26453}.  The optimized CUDA source
  is compiled into a CUBIN and disassembled to SASS.  A downstream SASS agent
  then searches for SASS-level optimizations, using \tool{} to encode the
  optimized instructions and rebuild candidate CUBINs.}
  \label{fig:cuda-sass-agent}
\end{minipage}
\vspace{0.04in}}
]

\begin{NoHyper}
\printAffiliationsAndNotice{}
\end{NoHyper}

\section{Introduction}

Performance-critical GPU kernels ultimately execute as NVIDIA
SASS machine code.  Source-level CUDA optimization determines algorithms, data
movement, tiling, and synchronization, while compiler lowering determines
register allocation, instruction scheduling, and machine-level control fields.
This boundary matters for workloads such as LLM training and inference, whose
core operators rely on specialized CUDA kernels.  FlashAttention demonstrates
the impact of source-level improvements through better work
partitioning, asynchrony, and low-precision support
~\cite{DBLP:journals/corr/abs-2205-14135,
DBLP:journals/corr/abs-2307-08691,DBLP:journals/corr/abs-2407-08608}.

Prior work demonstrates a second optimization opportunity below
PTX.  TuringAs enabled native SASS tuning and exposed optimizations unavailable
at higher levels~\cite{DBLP:conf/ppopp/YanWC20}.  CuAsmRL begins with an
optimized Triton kernel, reorders its SASS instructions under dependency
constraints, rebuilds the \cubin{}, and uses measured throughput as its reward.
On its Ampere LLM-kernel suite, it reports a 9\% mean improvement and
improvements of up to 26\%~\cite{DBLP:conf/cgo/HeY25}.
These gains show that even an optimized CUDA kernel can benefit
from additional SASS-level optimization after compilation.

Applying such SASS-level optimization systematically requires a
trustworthy way to reassemble a selected SASS candidate as an executable
\cubin{}.  A
\cubin{} stores GPU machine code in ELF sections.  NVIDIA's
\texttt{nvdisasm} recovers readable \sass{}, but NVIDIA provides no public tool
for assembling that \sass{} back into machine code on recent data-center
GPUs~\cite{nvidia-binary-utils-sm90}.  Manually maintained instruction-encoding
rules are costly to extend and may silently misencode newly encountered
instruction forms~\cite{maxas,DBLP:conf/ppopp/YanWC20}.

We present \tool{}, which learns exact 128-bit \sass{} encoders
from parsed disassembly paired with original \cubin{} instruction words.
\tool{} represents instruction encoding as vector-valued affine maps over
$\gf$.  It uses bitset Gaussian elimination to construct a compact basis and
emits an instruction word only when that basis uniquely determines the result.
\tool{} separates its learning algorithm from target-specific
control bits, relocation rules, and \cubin{} metadata.

Figure~\ref{fig:cuda-sass-agent} illustrates one application of
\tool{}.  A CUDA coding agent such as KernelPro first uses profiler feedback to
revise CUDA or CuTe source~\cite{DBLP:journals/corr/abs-2606-26453}.  A
downstream SASS agent can then search for additional improvements below PTX and
use \tool{} as its deterministic assembly backend.  \tool{} support-checks a
selected SASS candidate, emits its machine words, and rebuilds the \cubin{}
before external semantic testing and hardware measurement.

This paper makes five contributions:
\begin{enumerate}
  \setlength{\itemsep}{1pt}
  \item To our knowledge, the first formulation of learned
        \sass{} instruction encoding as vector-valued affine maps over $\gf$,
        with exact support checks.
  \item An incremental bitset Gaussian-elimination algorithm
        that detects inconsistent training data and retains a compact,
        mergeable basis instead of every instruction--word pair.
  \item A training pipeline that separates shared learning from
        target-specific machine-code rules, enabling extension across Hopper,
        Blackwell, Rubin, and future GPU targets.
  \item A continual-training procedure for extending a frozen
        encoder with additional \cubin{} observations, together with a tight
        lower bound on the independent observations required for a given set of
        unsupported queries.
  \item Large-scale empirical validation: F2Asm reconstructs
        3.41 billion executable bytes across Hopper, Blackwell, and Rubin
        exactly, while joint Blackwell training provides corpus-scale evidence
        that 5 SM targets share one SASS encoding scheme for common
        instructions.
\end{enumerate}

\section{Exact Bit-Linear Encoding}

\subsection{Forming the training system}
\label{sec:forming-training-system}

\begin{figure*}[t]
\centering
\begin{tikzpicture}[
    x=1cm,
    y=1cm,
    font=\sffamily\scriptsize,
    flow/.style={-{Latex[length=1.6mm]}, draw=navy, line width=0.52pt},
    panel/.style={draw=navy, rounded corners=2pt, line width=0.7pt,
                  fill=black!1},
    paneltitle/.style={font=\sffamily\scriptsize\bfseries, text=navy},
    mapbox/.style={rounded corners=1.7pt, line width=0.6pt, align=left,
                   inner xsep=5pt, inner ysep=3.5pt},
    contextbox/.style={draw=navy, fill=navy!4, rounded corners=1.7pt,
                       line width=0.6pt, align=center, minimum height=0.66cm,
                       inner xsep=5pt},
    cell/.style={draw=black!22, line width=0.35pt, minimum height=0.39cm,
                 minimum width=1.48cm, align=center,
                 inner xsep=1.5pt, inner ysep=0.5pt, outer sep=0pt,
                 text height=1.45ex, text depth=0.35ex},
    rowlabel/.style={font=\sffamily\tiny\bfseries, text=evalgray,
                     align=left}
  ]

  \node[draw=sassorange, fill=sassorange!6, rounded corners=2pt,
        line width=0.72pt, minimum width=16.62cm, minimum height=0.76cm,
        align=center, anchor=north] (source) at (8.40,5.55) {
    \textcolor{sassorange}{\textbf{SASS text}}\qquad
    \texttt{\textcolor{kernelproblue}{HMMA}%
      \textcolor{sassorange}{.16816.F32.BF16}
      R4, R68\textcolor{sassorange}{.reuse}, R72, R4}
  };

  \node[panel, minimum width=6.70cm, minimum height=3.55cm,
        anchor=north west] (leftpanel) at (0.00,4.35) {};
  \node[panel, minimum width=9.78cm, minimum height=3.55cm,
        anchor=north west] (rightpanel) at (7.02,4.35) {};

  \draw[flow] ([xshift=-5.05cm]source.south) -- (leftpanel.north);
  \draw[flow] ([xshift=3.51cm]source.south) -- (rightpanel.north);

  \node[paneltitle, anchor=west] at (0.25,4.07)
    {Form and modifier tuple};
  \node[paneltitle, anchor=west] at (7.27,4.07)
    {Parsed-value vector $v$ and derived operand signature $s$};

  \node[mapbox, draw=kernelproblue, fill=kernelproblue!6,
        text width=5.93cm, anchor=north] (formbox) at (3.35,3.74) {
    \textcolor{kernelproblue}{\textbf{Form}}\quad
    $f=\texttt{HMMA\_R\_R\_R\_R}$\\[-1pt]
    \textcolor{evalgray}{mnemonic $+$ ordered operand types $(D,A,B,C)$}
  };

  \node[mapbox, draw=sassorange, fill=sassorange!6,
        text width=5.93cm, anchor=north] (modbox) at (3.35,2.62) {
    \textcolor{sassorange}{\textbf{Modifier tuple}}\\[-1pt]
    \scalebox{0.90}{$m=(\texttt{HMMA},\texttt{16816},\texttt{F32},
      \texttt{BF16},\texttt{first-source reuse})$}\\[-1pt]
    \textcolor{evalgray}{mnemonic $+$ instruction- and operand-level modifiers}
  };

  \matrix (values) [
      matrix of nodes,
      anchor=north,
      row sep=-0.35pt,
      column sep=-0.35pt,
      nodes={cell},
      column 1/.style={nodes={cell, rowlabel, minimum width=1.42cm}},
      row 1/.style={nodes={cell, fill=black!5,
                           font=\sffamily\tiny\bfseries}},
      row 4/.style={nodes={cell, fill=kernelproblue!5}},
      row 5/.style={nodes={cell, fill=fasmgreen!5}},
      row 6/.style={nodes={cell, fill=fasmgreen!5}}
    ] at (11.84,3.82) {
      token & $\texttt{PT}^{\dagger}$ & \texttt{R4} &
              {\tiny\texttt{R68.reuse}} & \texttt{R72} & \texttt{R4} \\
      index $j$ & $0$ & $1$ & $2$ & $3$ & $4$ \\
      role & guard & $D$ & $A$ & $B$ & $C$ \\
      $v_j$ & $7$ & $4$ & $68$ & $72$ & $4$ \\
      sign & $+$ & $+$ & $+$ & $+$ & $+$ \\
      sentinel & {\tiny sentinel (7)} & {\tiny ordinary} &
                 {\tiny ordinary} & {\tiny ordinary} & {\tiny ordinary} \\
    };

  \coordinate (equalityrowcenter) at
    ($(values-6-1.south west)+(0,-0.195cm)$);
  \coordinate (equalityrowsoutheast) at
    ($(values-6-6.south east)+(0,-0.39cm)$);
  \path[draw=black!22, fill=fasmgreen!5, line width=0.35pt]
    (values-6-1.south west) rectangle (equalityrowsoutheast);

  \coordinate (eqone) at (values-1-3.center |- equalityrowcenter);
  \coordinate (eqfour) at (values-1-6.center |- equalityrowcenter);
  \node[font=\sffamily\tiny\bfseries, text=navy]
    at (values-1-1.center |- equalityrowcenter) {equality};
  \draw[draw=navy, line width=0.45pt] (eqone) -- (eqfour);
  \fill[navy] (eqone) circle (1.25pt);
  \fill[navy] (eqfour) circle (1.25pt);
  \node[font=\sffamily\tiny, text=navy, anchor=south east,
        xshift=-1.5pt, yshift=-2.5pt] at (eqone) {$1$};
  \node[font=\sffamily\tiny, text=navy, anchor=south west,
        xshift=1.5pt, yshift=-2.5pt] at (eqfour) {$4$};
  \node[font=\sffamily\tiny, text=navy, fill=fasmgreen!5,
        inner xsep=3pt, inner ysep=0.5pt]
    at ($(eqone)!0.5!(eqfour)$) {pair $(1,4)$: $v_1=v_4=4$};

  \draw[decorate, decoration={brace, amplitude=3pt},
        draw=fasmgreen, line width=0.6pt]
    ([xshift=0.05cm]values-5-6.north east) --
    ([xshift=0.05cm]equalityrowsoutheast)
    node[midway, right=4pt, text=fasmgreen,
         font=\sffamily\scriptsize\bfseries] {$s$};

  \draw[draw=evalgray, dashed, rounded corners=1pt, line width=0.45pt]
    ([xshift=0.4pt,yshift=0.4pt]values-1-2.south west) rectangle
    ([xshift=-0.4pt,yshift=-0.4pt]values-1-2.north east);

  \node[contextbox, text width=3.15cm, anchor=west] (pkey) at (0.82,-0.13) {
    $p=(m,s)$\\[-1pt]
    \textcolor{evalgray}{\tiny\texttt{InstructionContext}}
  };
  \node[contextbox, text width=4.25cm, right=0.92cm of pkey] (ckey) {
    $c=(f,p)=(f,m,s)$\\[-1pt]
    \textcolor{evalgray}{\tiny exact context}
  };
  \node[contextbox, draw=fasmgreen, fill=fasmgreen!6,
        text width=4.15cm, right=0.92cm of ckey] (training) {
    group observations with identical $c$\\[-1pt]
    \textcolor{evalgray}{\tiny Gaussian elimination within the group}
  };

  \draw[flow] (pkey) -- (ckey);
  \draw[flow] (ckey) -- (training);

  \node[font=\sffamily\tiny, text=evalgray, anchor=north west,
        text width=16.55cm, align=left]
    at (0.03,-0.57) {
      $j$ is the zero-based index in $v$, beginning with the guard; the
      mnemonic and modifiers are not indexed.\\[-1pt]
      $\dagger$ parser-inserted implicit guard; $+$ denotes nonnegative and
      $-$ denotes negative; sentinel classes are operand-type-specific.
    };
\end{tikzpicture}
\caption{Instruction-context anatomy for the HMMA example.  The parser produces
$f$, $m$, $v$, and $s$.  The operand signature $s$ comprises
the sign, sentinel, and equality rows.  Then $p=(m,s)$ and $c=(f,p)$ define
the initial exact-context training group.}
\label{fig:encoding-context-anatomy}
\end{figure*}
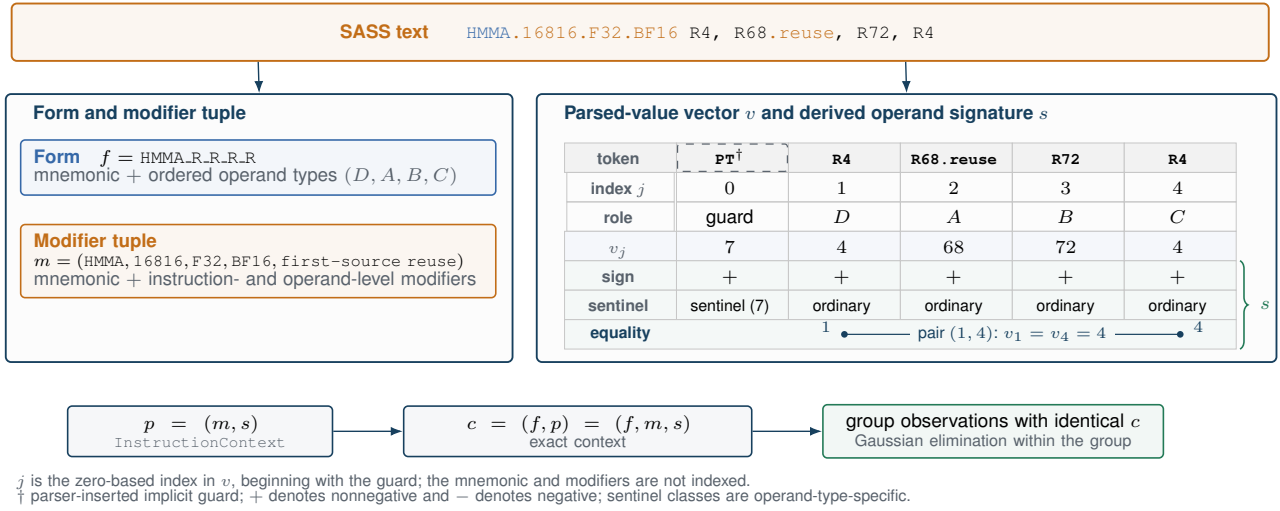

We represent instruction encoding over the binary field
$\gf=\{0,1\}$, where addition is XOR and multiplication is
AND~\cite{DBLP:books/crc/13/MP2013}.
A form $f$ combines a mnemonic with ordered operand types, such
as the Tensor Core \texttt{HMMA\_R\_R\_R\_R}.
Let $m$ be the ordered tuple containing the mnemonic followed by
all parsed instruction- and operand-level modifiers.
Let $v$ be the \emph{parsed-value vector} that F2Asm bit-vectorizes to form
the feature vector $\phi(x)$ defined below.  The parser places the instruction
guard first, followed by operand values in textual order.  Here the implicit
\texttt{PT} guard maps to 7, while \texttt{R4}, \texttt{R68.reuse},
\texttt{R72}, and \texttt{R4} contribute $4,68,72,4$, giving
$v=(7,4,68,72,4)$; the \texttt{reuse} modifier remains in $m$.  Let $s$ be
the \emph{operand signature}: the signs of the parsed values, their
operand-type-specific sentinel classes, and the pairs of positions containing
equal values.  F2Asm uses $s$ to distinguish cases that may require different
affine maps, then tests whether those cases can share one.  Here all values are
nonnegative, only $v_0=7$ is the \texttt{PT} sentinel (all others are
ordinary), and $v_1=v_4=4$, giving the equality pair $(1,4)$.  The
implementation represents $p=(m,s)$ as an instance of
the \texttt{InstructionContext} class.  The \emph{exact context} $c=(f,p)$ groups
training observations with identical $f$, $m$, and
$s$.  Figure~\ref{fig:encoding-context-anatomy} shows the
resulting $f$, $m$, $v$, $s$, $p$, and $c$ for one HMMA instruction; in
particular, \texttt{R68.reuse} contributes 68 to $v$ and
\texttt{first-source reuse} to $m$ (see also
Appendix~\ref{app:instruction-lifecycle}).
For each exact-context group, F2Asm applies Gaussian elimination to the
training system
$\Phi W=Y$ (Eq.~\ref{eq:training-system} and
Section~\ref{sec:streaming-elimination}) and retains a basis of linearly
independent rows.  This preserves the system's solution set.

Each training observation $i$ is a pair $(x_i,y_i)$, where
$x_i$ is the textual \sass{} instruction produced by \texttt{nvdisasm} and
parsed into $f_i$, $m_i$, and $v_i$, and $y_i\in\gf^{128}$ is its original
instruction word with
scheduler-control bits cleared to zero.  \tool{} handles scheduler control
separately using the target-specific bit layout
(Appendix~\ref{app:instruction-lifecycle}).  For the one \texttt{DEPBAR.LE}
context where \texttt{nvdisasm} omits bits 122 and 124, \tool{} also clears
these bits during training and preserves their original values separately for
byte-exact reconstruction (Appendix~\ref{app:depbar-opaque}).

\begingroup
The signs, sentinel classes, and equalities derived from $v_i$
form operand signature $s_i$, so $p_i=(m_i,s_i)$ and $c_i=(f_i,p_i)$.
F2Asm constructs the feature
vector
{\small
\begin{equation}
\phi(x_i)=[1\parallel\operatorname{bits}_{64}(v_{i,0})\parallel\cdots
\parallel\operatorname{bits}_{64}(v_{i,k-1})]\in\mathbb F_2^{1+64k}.
\label{eq:parsed-value-feature-map}
\end{equation}}

The feature vector $\phi(x_i)$ supplies the numerical input to an affine map.
F2Asm first trains each exact $(f,m,s)$ group separately, then
tests whether groups with the same $(f,m)$ can share a map.  It trains affine
maps independently for each $(f,m)$ because $\phi(x)$ contains only $v$.
Since $s$ is derived from $v$, groups with different observed $s$ may still
follow the same affine map.  F2Asm combines their bases and applies Gaussian
elimination: if the equations are consistent and the output masks match, it
stores one shared map; otherwise, it retains separate maps selected by $s$.
The following SM107 case illustrates how a modifier can change learned-map
assignment without changing the feature vector: \texttt{nvdisasm} omits
\texttt{.PHASE1} from
\texttt{WARPSYNC.}\allowbreak\texttt{COLLECTIVE R88, 0x9fa0} even when bit 18
of the original word is set.  \tool{} restores the modifier before parsing.
This leaves $f_i$, $v_i$, and $\phi(x_i)$ unchanged but changes the modifier
tuple in $p_i$, thereby selecting the map trained for \texttt{.PHASE1}
observations, which emits bit 18 as 1.

\noindent\begin{minipage}{\columnwidth}
When training a shared encoder, \tool{} pools observations from all SM targets
assigned to the same learned map.  The SM target is neither a feature nor a map
selector.  The derivation below considers one learned map, so we write $\Phi$,
$Y$, and $W$ without a map subscript.
\end{minipage}
\par
\endgroup

For observation $i$, the \emph{encoding equation} is
\begin{equation}
  \underbrace{y_i}_{1\times128}
  = \underbrace{\phi(x_i)}_{1\times d}
    \underbrace{W}_{d\times128}
  = \bigoplus_{j:\phi_j(x_i)=1}
    \underbrace{w_j}_{1\times128}.
  \label{eq:encoding-model}
\end{equation}

Here $w_j$ is row $j$ of $W$.  The symbol $\bigoplus$ denotes
XOR-reduction: it selects the rows whose feature bits equal 1 and XORs them.
This is ordinary row-vector--matrix multiplication over $\gf$, where scalar
multiplication is AND and addition is XOR.

Observations assigned to the same learned map are stacked to
form one training system.
\begin{equation}
  \underbrace{\Phi}_{n\times d}
  \underbrace{W}_{d\times128}
  = \underbrace{Y}_{n\times128}
  \quad\text{over }\gf.
  \label{eq:training-system}
\end{equation}
Row $i$ of $\Phi$ is $\phi(x_i)$, and row $i$ of $Y$ is
$y_i$.  The dense matrix $W$ is convenient notation for these constraints;
the implementation does not need to materialize an arbitrary solution outside
the observed feature span.

For the HMMA example above, the shared-map key contains form
$f=\texttt{HMMA\_R\_R\_R\_R}$ together with the modifier tuple $m$, while
$c=(f,p)$ is its exact encoding context.
The leading value 7
encodes \texttt{PT},
NVIDIA's always-true predicate: the 3-bit predicate field assigns 0--6 to
programmable registers \texttt{P0}--\texttt{P6} and $7=111_2$ to
\texttt{PT}.  Let $\parallel$ denote vector concatenation, and let
$\operatorname{bits}_{64}(a)$ list the 64 bits of $a$ from least to most
significant.  The resulting \emph{SM100 HMMA feature vector} is
\begin{equation}
\begin{aligned}
\phi(x)=[1
 &\parallel \operatorname{bits}_{64}(7)
 \parallel \operatorname{bits}_{64}(4)
 \parallel \operatorname{bits}_{64}(68)\\
 &\parallel \operatorname{bits}_{64}(72)
 \parallel \operatorname{bits}_{64}(4)]
 \in\gf^{321}.
\end{aligned}
\label{eq:hmma-feature-vector}
\end{equation}
The leading $1$ supplies the affine bias (constant term), while
the five 64-bit subvectors are the little-endian binary encodings of
\texttt{PT}, destination $D$, source $A$, source $B$, and accumulator $C$, as
labeled in Figure~\ref{fig:encoding-context-anatomy}.
Appendix~\ref{app:instruction-lifecycle} follows this exact instruction through training and assembly.

\subsection{Training with streaming Gaussian elimination}
\label{sec:streaming-elimination}

\tool{} performs Gaussian elimination incrementally, processing
one training observation at a time.  It immediately reduces each augmented row
against the retained pivot basis and stores the row only if it increases the rank.
Memory therefore scales with the retained basis rather than the observation
count.  This solves Eq.~\ref{eq:training-system} on the observed span without
materializing a dense $W$.  Each streamed row is
\begin{equation}
  \underbrace{[\,r\mid z\,]}_{1\times(d+128)}
  =
  [\,\underbrace{\phi(x_i)}_{1\times d}
  \mid \underbrace{y_i}_{1\times128}\,]
  \quad\text{over }\gf
  \label{eq:augmented-row}
\end{equation}
and maintains a reduced basis
$B=\{(r_k,z_k)\}$.  Each $r_k$ is a retained feature equation and $z_k$ is
the 128-bit output obtained by applying the same row operations to $Y$.  The
rank is the number of linearly independent rows retained after elimination.

For each row $(r,z)$, training (1) XOR-reduces it against the
stored pivots; (2) discards zero input and output residuals; (3) reports a
contradiction for a zero input but nonzero output residual; otherwise, (4)
selects the highest set input bit as a pivot, (5) clears that pivot from stored
rows, and inserts $(r,z)$.
Feature rows and output words are packed as integers, so XOR
implements row addition directly.  Training processes every observation but
stores only the linearly independent basis rows.  A fixed observation order
produces a deterministic basis.  Bases computed from separate corpus shards
can be merged by eliminating their retained rows, enabling parallel collection
without storing the full corpus.

An \emph{encoder} maps parsed SASS to 128-bit words; its
maps, bases, and selection metadata form the \emph{model repository}.

\subsection{SASS encoding with the trained model}

\revisioncolor{The trained model is supported on the observed feature subspace}
\begin{equation}
  S=\operatorname{rowspan}(\Phi)\subseteq\gf^d.
  \label{eq:evidence-span}
\end{equation}
Here, $\Phi\in\gf^{n\times d}$ is the training feature matrix;
its $i$th row is the feature vector $\phi(x_i)$ of observation $i$.  Its row
span is
\begin{equation}
  \revisioncolor{\operatorname{rowspan}(\Phi)
  =\{\,a^\top\Phi:a\in\gf^n\,\}.}
  \label{eq:rowspan-definition}
\end{equation}
Because arithmetic is over $\gf$, this is the set of all XOR
combinations of the observed feature rows.
\revisioncolor{Therefore, a query instruction $x$ is supported exactly when
$\phi(x)\in S$, meaning that $\phi(x)$ can be expressed as an XOR of observed
feature rows.}
The observations are consistent exactly when
\begin{equation}
  \operatorname{rank}([\,\Phi\mid Y\,])
  =\operatorname{rank}(\Phi).
  \label{eq:training-consistency}
\end{equation}
When this condition holds, elimination defines one unique linear map
\begin{equation}
  T:S\rightarrow\gf^{128},
  \qquad T(\phi(x_i))=y_i.
  \label{eq:evidence-map}
\end{equation}
The stored basis pairs are the compact representation of this
map: $T(r_k)=z_k$.  A dense $W$ in Eq.~\ref{eq:training-system} is any linear
extension of $T$ to the ambient space $\gf^d$; such an extension need not be
unique outside $S$.  Therefore, exact training requires consistency, not
$\operatorname{rank}(\Phi)=d$.  Appendix~\ref{app:streaming-elimination-example} illustrates training and encoding
with a worked example.

To encode a query instruction $x$, \tool{} forms
$q=\phi(x)$ and reduces $q$ with the stored $r_k$ rows.  Whenever it XORs
$r_k$ from the query, it also XORs $z_k$ into an initially zero 128-bit
accumulator.  A zero residual proves $q\in S$; the accumulator is then the
unique supported value $T(q)$.  A nonzero residual proves $q\notin S$.
\tool{} emits an instruction word only when the query is supported by the
retained basis; otherwise, it reports an unsupported instruction and emits no
output.
Although \tool{} rejects unsupported inputs,
Section~\ref{sec:roundtrip-experiments} shows that this safeguard did not limit
coverage: every instruction encountered across 3,225 production \cubin{}s was
encoded successfully.

\subsection{Continual training of a trained encoder}
\label{sec:continual-extension}

Continual training is particularly useful for newly
introduced GPU architectures such as Rubin, where the available \cubin{}
training corpus is initially limited but grows with subsequent library and
software releases.

Let $g$ identify a map group: $g=(f,m)$ for a shared map or
$g=(f,m,s)$ for a signature-specific map.  When new CUBINs expose unsupported
queries, their training rows form $E$, partitioned by map group into $E_g$.

\begingroup
\setlength{\intextsep}{4pt}
\begin{algorithm}[H]
\caption{Continual update of a frozen encoder}
\label{alg:continual-update}
\begingroup
\normalsize
\setboolean{ALC@noend}{true}
\begin{algorithmic}[1]
\REQUIRE Model repository $R$; training-row batch $E$
\STATE Verify $E$; initialize $R'$ from $R$ without modifying $R$
\FOR{each affected map group $g$}
  \STATE Set $B'_g$ to its retained basis, or $\varnothing$ for a new map group
  \FOR{each $(r,z)\in E_g$}
    \STATE Reduce $(r,z)$ against $B'_g$ to $(\bar r,\bar z)$
    \STATE \textbf{if} $\bar r\neq 0$ \textbf{then} insert the pivot
    \STATE \textbf{else if} $\bar z\neq 0$ \textbf{then} abort
    \STATE \textbf{else} record a redundant consistency check
  \ENDFOR
\ENDFOR
\STATE \textbf{return} the updated model repository $R'$
\end{algorithmic}
\endgroup
\end{algorithm}
\endgroup

If a new observation contradicts a shared $(f,m)$ map,
Algorithm~\ref{alg:continual-update} aborts the update at line~7.  Automatic
repartitioning would split that map into signature-specific $(f,m,s)$ groups
and rebuild the affected maps.  The experiment in
Section~\ref{sec:rubin-continual} did not require repartitioning, so we leave it
to future work.

For each map group $g$, let $B_g$ be the retained feature
basis and $Q_g=\{\phi(x):x\text{ is unsupported in }g\}$ the unsupported-query
feature rows.  Set
$S_g=\operatorname{span}(B_g)$ and
$S_g^+=\operatorname{span}(B_g\cup Q_g)$.  Then
{\small
\begin{equation}
  \delta_g=\operatorname{rank}(B_g\cup Q_g)
  -\operatorname{rank}(B_g)
  =\dim(S_g^+/S_g).
  \label{eq:continual-rank-bound}
\end{equation}
}
By the quotient-dimension identity~\cite{Hungerford1980}, each
rank-increasing observation adds one dimension.  Thus any sufficient observation
set for group $g$ must contain at least $\delta_g$ observations, and greedy insertion
retains exactly $\delta_g$~\cite{Oxley2011}; therefore, the repository-wide
minimum is $\delta=\sum_g\delta_g$.  When an instruction query is unsupported,
an autonomous SASS agent can invoke Algorithm~\ref{alg:continual-update} with
new CUBIN observations to update the affected map groups.

\section{Training Data and Pipeline}

The training data are instruction--encoding pairs extracted after SHA-256
deduplication.  The SM90/90a corpus comes
from pinned NVIDIA and third-party production libraries; SM100 comes from
NVIDIA CUDA 13.3 packages; and SM107 comes from NVIDIA CUDA 13.4
Developer Preview archives~\cite{nvidia-cuda-13-4-preview-archives}.
Table~\ref{tab:evidence} summarizes these corpora and their
reduction into training bases.  We call a CUBIN
\emph{authenticated} when its provenance is recorded, its SHA-256 digest and
target metadata are verified, and its derived observations remain traceable to
the source artifact.  All CUBINs in this paper are authenticated.

The training-data pipeline, which we call the collector, has
three checkpointed stages.
Stage 1 verifies each indexed CUBIN and stores
compressed \texttt{nvdisasm -hex -c} output in a content-addressed shard.
Stage 2 parses each verified shard into traceable
instruction-observation records.
Stage 3 replays those records in corpus order through streaming
Gaussian elimination and atomically writes the retained rows, per-CUBIN
ledger, and evidence manifest.

The pipeline is designed for automated target onboarding and
can be driven by an autonomous coding agent.  Its stages are deterministic,
non-interactive, checkpointed, and emit machine-readable pass/fail manifests.
An architecture profile specifies the instruction layout, control fields,
relocation and metadata rules, parser recovery, and model repository for a
compatible group of SM targets.  Compatible targets may share one profile and
encoder.  To onboard a new \texttt{sm\_*} target, an agent defines or extends
an architecture profile, constructs a verified corpus index, pins the
disassembler, runs model training and round-trip tests, and uses manifest
failures to guide the next iteration.

\texttt{nvdisasm} sometimes hides distinctions present in
machine code: different 128-bit instruction words may disassemble to the same
SASS text.  \tool{} examines the original instruction bits to recover those
distinctions; if it cannot do so safely, it rejects the observation.  For
example, in a Rubin SM107 cuSPARSE CUBIN, \texttt{nvdisasm}
omits the \texttt{.PHASE1} modifier from
\texttt{WARPSYNC.}\allowbreak\texttt{COLLECTIVE R88, 0x9fa0}, although bit 18
of the original word encodes it.  The collector restores \texttt{.PHASE1}
before training.

\revisioncolor{Across the corpora in Table~\ref{tab:evidence}, the collector
reduces 212,937,948 accepted instruction observations to 1,063,639 basis rows
by applying streaming Gaussian elimination separately within each exact
$(f,p)$ context, reducing the row count by approximately $200\times$.  \tool{}
then trains the encoding models using the procedure in
Section~\ref{sec:forming-training-system}.  The model repository's final map
and basis counts after pooling appear in Table~\ref{tab:repository-structure}.}

\begin{table*}[t]
\centering
\caption{\revisioncolor{Training corpora and exact-context
evidence compaction.}}
\label{tab:evidence}
\scriptsize
\setlength{\tabcolsep}{3pt}
\begin{tabular*}{\textwidth}{@{\extracolsep{\fill}}l l p{0.34\textwidth} rrrrr@{}}
\toprule
Target & GPU architecture & Corpus sources (CUBINs) & Total & Raw rows &
\revisioncolor{Exact-context basis rows} &
\revisioncolor{Exact contexts} & Skips\textsuperscript{a} \\
\midrule
SM90/90a & Hopper & NVIDIA TensorRT-LLM (1,108); NVIDIA Transformer Engine (60);
  vLLM (239); FlashAttention (72); SGLang (60); xFormers (8);
  bitsandbytes (3); rVLLM (1)
  & 1,551 & 169.87M & 440,308 & 60,412 & 200 \\
SM100 & Blackwell & NVIDIA cuBLAS (378); cuDNN (283); cuSOLVER (194); NPP (182);
  cuSPARSE (100); nvJPEG (10); cuRAND (7); cuSPARSELt (3)
  & 1,157 & 25.09M & 385,141 & 59,894 & 0 \\
SM107 & Rubin & NVIDIA cuSOLVER (203); NPP (197); cuSPARSE (100); nvJPEG (10);
  cuRAND (7)
  & 517 & 17.98M & 238,190 & 37,277 & 13,524 \\
\midrule
\multicolumn{3}{@{}l}{Total} & 3,225 & 212,937,948 & 1,063,639 & 157,583 & 13,724 \\
\bottomrule
\end{tabular*}
\parbox{0.98\textwidth}{\vspace{2pt}{\fontsize{6}{6.6}\selectfont
\raisebox{0.25ex}{\textit{a}} Skips are symbolic-address rows
whose \texttt{nvdisasm} operand could not be matched unambiguously to its ELF
relocation, preventing safe separation of instruction bits from
linker-controlled bits.  They were excluded from training; other
rows and relocation rules retained opcode coverage.  Unexpected skips were
zero, and all CUBINs passed byte-exact round trip.  SM90/90a:
172 \texttt{MOV}
symbol-plus-\texttt{@srel} expressions and 28 \texttt{UMOV} symbol mismatches
across nine CUBINs.  SM107: 13,506 cross-section \texttt{CALL}s and 18
\texttt{UMOV} relocation expressions.}}
\end{table*}

\begin{table}[t]
\centering
\caption{\revisioncolor{Final encoding-repository structure.  Each $(f,m)$
group stores one shared affine map, or one per operand signature when sharing
fails; basis rows are summed across maps.}}
\label{tab:repository-structure}
\scriptsize
\setlength{\tabcolsep}{4pt}
\revisioncolor{%
\begin{tabular*}{\columnwidth}{@{\extracolsep{\fill}}lrrr@{}}
\toprule
Target & $(f,m)$ groups & Affine maps & Final basis rows \\
\midrule
SM90/90a & 2,865 & 2,936 & 50,099 \\
SM100    & 3,669 & 3,673 & 60,159 \\
SM107    & 2,896 & 2,896 & 44,882 \\
\midrule
Total    & 9,430 & 9,505 & 155,140 \\
\bottomrule
\end{tabular*}}
\end{table}

\section{Experiments}
\label{sec:roundtrip-experiments}

The experiments evaluate four properties of \tool{}.
First, byte-exact round-trip testing validates learned encodings across Hopper,
Blackwell, and Rubin corpora.  Second, joint training tests one encoder across
5 Blackwell SM targets.  Third, controlled held-out and unseen-library
evaluations characterize one encoder shared by 3 Rubin SM targets.  Fourth,
continual training extends the frozen Rubin encoder using newly available
CUBINs.

\subsection{Round-trip evaluation}

Round-trip testing is \tool{}'s end-to-end acceptance test
because an assembler must reproduce machine code exactly.  Earlier CUDA
assemblers also used exact reconstruction for validation: MaxAs reported
zero-error disassembly--reassembly of \texttt{cublas\_device.lib}
~\cite{maxas}; Decoding CUDA Binary compared generated benchmark code with the
original~\cite{DBLP:conf/cgo/HayesHHCZ19}; and CuAssembler compared re-encoded
instruction words with their original encodings~\cite{cuassembler}.  \tool{}
extends this criterion to deduplicated production CUBINs with
complete executable-section accounting.
The round-trip test preserves duplicate section names by matching
sections by name and occurrence index in ELF section-header order, then checks
every matched section byte-for-byte.  For each deduplicated, text-bearing
CUBIN, \tool{} disassembles the
executable code, parses the SASS, re-encodes every instruction, rebuilds the
CUBIN, and compares every \texttt{.text.*} section byte-for-byte with the
original.  A CUBIN passes only when all executable sections are present and
identical.

Table~\ref{tab:results} reports the round-trip results.
All 3,225 text-bearing CUBINs passed: \tool{} reconstructed
150,308 executable sections
(3,407,226,752 bytes) byte-for-byte, with 0 failures and 0 errors.  On
Rubin SM107, the latest GPU architecture supported by \tool{}, all 517 CUDA
13.4 Developer Preview CUBINs passed, covering
32,676 executable sections and 279
instruction forms.  This includes the
\texttt{WARPSYNC.}\allowbreak\texttt{COLLECTIVE.PHASE1} case described in
Section~3, where \texttt{nvdisasm} omits an encoding-significant modifier that
\tool{} recovers from the original instruction word.

The SM100 evaluation identifies three required disassembly
normalizations and one CUBIN-format requirement.  The three normalizations---
exact QNaN payload recovery, symbolic \texttt{RZ}/\texttt{URZ}
versus literal \texttt{R255}/\texttt{UR63}, and PC-relative \texttt{WARPSYNC}
targets---reduce $\gf$ conflicts from 5,830 to zero over the same 25,089,192
observations.  The SM100 corpus contains a parallel per-kernel
ELF namespace: all 30,491 \texttt{.text.*} sections have matching
\texttt{.nv.capmerc.text.*} sections, accompanied by \texttt{.nv.merc.*}
metadata, relocations, symbols, and debug data.  This structure is consistent
with a Blackwell finalization capsule.\footnote{Independent reverse-engineering
analysis describes this format as Capsule Mercury:
\url{https://gh.evko.io/nvopen-tools/ptxas/codegen/capmerc.html}.}  \tool{}
preserves it without relying on its undocumented semantics.

\revisioncolor{SM107's exact-context evidence systems are strongly low-rank:
17,977,652 \newrevisioncolor{admitted observations} reduce to 238,190 exact-context basis rows;
the median exact-context rank is 3 and 95\% are at most 21.  Repository-level
pooling further reduces these to 44,882 stored basis rows across 2,896 $(f,m)$
groups, all represented by shared maps.}
The SM107 repository contains 23 form signatures absent from
SM100.  \texttt{IADD}, \texttt{IMNMX}, and \texttt{VISET} do not appear in
either the SM90 or SM100 repository and account for 1,437,961 SM107
\newrevisioncolor{instruction observations} (8.00\%).

\begin{table}[t]
\centering
\caption{Round-trip results for the 3,225-CUBIN release corpus.
All executable text sections matched their originals byte-for-byte.
P/F/E denotes passed/failed/errors.}
\label{tab:results}
\scriptsize
\begin{tabular}{@{}lrrrrr@{}}
\toprule
Target & Forms & CUBINs & Sections & Text MiB & P/F/E \\
\midrule
SM90/90a & 389 & 1,551 & 87,141 & 2,592.0 & 1,551/0/0 \\
SM100    & 326 & 1,157 & 30,491 &   382.8 & 1,157/0/0 \\
SM107    & 279 &   517 & 32,676 &
  274.5 &   517/0/0 \\
\bottomrule
\end{tabular}
\end{table}

\subsection{One SASS encoder for 5 Blackwell SM targets}

\tool{} reveals that SM100, SM100f, SM100a, SM103, and SM103a
use the same \revisioncolor{instruction-encoding scheme} for their common instruction forms and
contexts.  To test whether these targets require distinct learned encoding
rules, we pooled 18,015 CUBINs into one training corpus.  Each observation
retains its SM target only for corpus verification, target-specific
instruction legality checks, and CUBIN reconstruction; it does not affect the
learned encoder.  The \texttt{flashinfer-cubin==0.6.13}
package~\cite{flashinfer-cubin-0613} supplies 13,646 of the 14,036 SM100f
CUBINs and all 1,952 SM103a CUBINs.

\begin{table}[t]
\centering
\caption{Shared-encoder round-trip results for 5 Blackwell SM
targets.  \emph{Coverage}: full corpus or downsampled test percentage.
P/F/E denotes passed/failed/errors.}
\label{tab:blackwell-shared}
\setlength{\tabcolsep}{2pt}
\scriptsize
\sbox{\blackwellsharedtablebox}{%
\begin{tabular}{@{}lrrlrrr@{}}
\toprule
SM & Train & Test & Coverage & Sect. & MiB & P/F/E \\
\midrule
SM100\textsuperscript{a} & 1,217 & 1,217 & Full
  & 30,611 & 384.7 & 1,217/0/0 \\
SM100f & 14,036 & 2,808 & 20.0\% & 19,840 & 505.7 & 2,808/0/0 \\
SM100a &    314 &   314 & Full                  &    636 &  20.2 &   314/0/0 \\
SM103  &    496 &   496 & Full                  & 20,014 & 256.5 &   496/0/0 \\
SM103a &  1,952 &   500 & 25.6\% &  1,000 &  29.6 &   500/0/0 \\
\midrule
\textbf{Total} & \textbf{18,015} & \textbf{5,335} & Mixed &
  \textbf{72,101} & \textbf{1,196.7} & \textbf{5,335/0/0} \\
\bottomrule
\end{tabular}}
\begin{minipage}{\wd\blackwellsharedtablebox}
  \noindent\usebox{\blackwellsharedtablebox}\par
  \vspace{2pt}
  \raggedright
  \fontsize{6}{6.6}\selectfont
  \raisebox{0.25ex}{\textit{a}} Includes the 1,157 SM100 CUBINs
  in Table~\ref{tab:evidence} plus 60 from
  \texttt{flashinfer-cubin==0.6.13}.\par
\end{minipage}
\end{table}

Training processed 127,926,344 instruction observations across 388 forms and
77,926 encoding contexts.  Of these, 49,566 occurred in both the SM100-variant
and SM103-variant corpora.  The remaining 28,360 occurred in only one variant
group but were also included in the shared encoder.
The five target-specific collection runs retained 1,036,618
basis rows across their exact-context systems.  Joint training pooled and
re-eliminated these rows across compatible contexts, leaving 75,692 basis rows
\revisioncolor{across the shared encoder's $(f,m)$ groups.}
\revisioncolor{The shared encoder contains 4,397 $(f,m)$ groups.  Of
these, 4,396 use one map shared across all operand signatures $s$.  Only one
\texttt{DEPBAR} $(f,m)$ group requires exact-signature fallback, using five maps,
for 4,401 maps in total.}
All training observations were consistent with one shared encoder; none required
a target-specific learned map or output rule.

Table~\ref{tab:blackwell-shared} reports byte-exact round-trip
results for the shared encoder artifact.  We tested the full SM100, SM100a, and
SM103 corpora; for SM100f and SM103a, we deterministically downsampled the test
sets while covering every observed encoding context.  All
5,335 CUBINs passed, covering 72,101 executable sections and 1,254,796,416
bytes.  Together, these results empirically establish that SM100, SM100f,
SM100a, SM103, and SM103a share the same SASS instruction-encoding scheme.  To
our knowledge, \tool{} provides the first corpus-scale demonstration of this
shared encoding across 5 Blackwell SM targets.

\revisioncolor{\subsection{One SASS encoder for 3 Rubin SM targets}}

\revisioncolor{We trained one encoder jointly for \texttt{sm\_107}, \texttt{sm\_107a}, and
\texttt{sm\_107f}. We assembled a Rubin development corpus of 18,578
CUBINs: 517 from CUDA 13.4 libraries, 17,999 from FlashInfer, 52 compiled from
CUTLASS, and 10 targeted probes. We divided the corpus into 14,970 training
CUBINs and 3,608 held-out CUBINs. CUBINs sharing a source or build group,
executable payload, or target-sibling relationship were kept in the same
partition. Training comprised 510 \texttt{sm\_107}, 14,431 \texttt{sm\_107a},
and all 29 \texttt{sm\_107f} CUBINs; the held-out set comprised 10
\texttt{sm\_107} and 3,598 \texttt{sm\_107a} CUBINs. The held-out CUBINs were
excluded from training while ensuring that every normalized instruction query
they contain was covered by the training set. No independent
\texttt{sm\_107f} split satisfied the query-coverage requirement.}

\revisioncolor{Training processed 71,927,724 instruction observations and retained 296,403
exact-context basis rows. Repository training pooled these rows into 3,428
$(f,m)$ groups, each represented by one shared affine map, with 54,957 final
basis rows across 368 instruction forms.}

\begin{table*}[t]
\centering
\caption{\revisioncolor{Rubin development-corpus partition and round-trip
results for the trained encoder. P/F/E denotes passed/failed/error CUBINs.}}
\label{tab:rubin-development}
\scriptsize
\setlength{\tabcolsep}{6pt}
\revisioncolor{%
\begin{tabular}{@{}lrrrrr@{}}
\toprule
Partition & CUBINs & Instructions & Unique queries & Text sections & P/F/E \\
\midrule
Training & 14,970 & 71,927,724 & 4,412,203 & 59,921 & 14,970/0/0 \\
Held-out & 3,608 & 12,788,800 & 740,247 & 9,017 & 3,608/0/0 \\
\bottomrule
\end{tabular}}
\end{table*}

\revisioncolor{As Table~\ref{tab:rubin-development} shows, all training and held-out CUBINs
passed strict round-trip using the frozen encoder and toolchain, and every
executable section matched exactly.}

\revisioncolor{To evaluate the frozen encoder beyond this development corpus,
we used CUBINs from NVIDIA cuTile Python~\cite{nvidia-cutile-python} and
GROMACS~\cite{DBLP:conf/easc/PallAKHL14} as unseen-library test sets.
Table~\ref{tab:rubin-unseen} reports frozen-encoder coverage on these CUBINs.
Unsupported queries were rejected without encoding, and fully supported CUBINs
were required to pass strict round-trip. Section~\ref{sec:rubin-continual} uses
the unsupported queries from this evaluation to test continual adaptation.}

\begin{table*}[t]
\centering
\caption{\revisioncolor{Pre-adaptation coverage of the trained Rubin encoder on
unseen-library CUBINs from NVIDIA cuTile Python and GROMACS. Supported counts
instruction occurrences accepted by the encoder; coverage is supported
occurrences divided by instructions.}}
\label{tab:rubin-unseen}
\scriptsize
\setlength{\tabcolsep}{5pt}
\newsavebox{\rubinunseentabular}
\savebox{\rubinunseentabular}{\revisioncolor{%
\begin{tabular}{@{}llrrrrrr@{}}
\toprule
Source & Target & Text CUBINs & Instructions & Supported & Coverage & Fully supported & Round-trip \\
\midrule
cuTile & \texttt{sm\_107a} & 5 & 20,040 & 16,338 & 81.5\% & 0 & N/A \\
GROMACS & \texttt{sm\_107} & 23 & 364,800 & 311,666 & 85.4\% & 4 & 4/4 \\
\bottomrule
\end{tabular}}}
\usebox{\rubinunseentabular}
\parbox{\wd\rubinunseentabular}{\vspace{2pt}{\fontsize{6}{6.6}\selectfont
\revisioncolor{Sources were fixed before payload inspection: NVIDIA cuTile
Python commit \texttt{068039e} and GROMACS v2026.3.}}}
\end{table*}

\Needspace{7\baselineskip}
\subsection{\revisioncolor{Continual adaptation of the Rubin encoder}}
\label{sec:rubin-continual}

\revisioncolor{For a newly introduced architecture such as Rubin, initially
available CUBIN corpora can be large yet still sample the instruction domain
unevenly.  Continual training lets \tool{} extend an existing encoder as
additional CUBINs become available.  We used Algorithm~\ref{alg:continual-update}
to update the Rubin encoder from the unsupported queries found in the cuTile
and GROMACS evaluation.  The 24 CUBINs containing
unsupported queries became adaptation data; the four fully supported GROMACS
CUBINs were not needed for the update.}

\begin{table}[t]
\centering
\caption{\revisioncolor{Unsupported Rubin queries before adaptation and the
minimum number of additional independent instruction observations required to
support them.  Obs. denotes unsupported instruction observations.}}
\label{tab:rubin-continual}
\scriptsize
\setlength{\tabcolsep}{2.5pt}
\revisioncolor{%
\begin{tabular*}{\columnwidth}{@{\extracolsep{\fill}}lrrrrr@{}}
\toprule
Source & Obs. & Queries & Forms & Maps & $\sum_g\delta_g$ \\
\midrule
cuTile   & 3,702  & 3,421  & 21 & 33  & 104 \\
GROMACS  & 53,134 & 13,738 & 63 & 173 & 1,400 \\
\midrule
Combined & 56,836 & 17,159 & 75 & 206 & 1,504 \\
\bottomrule
\end{tabular*}}
\end{table}

\revisioncolor{Across the affected map groups,
Algorithm~\ref{alg:continual-update} processed 5,431 evidence rows.  Of these,
1,504 were linearly independent and expanded the encoder, matching the bound
in Table~\ref{tab:rubin-continual}; the remaining 3,927 confirmed consistency
with the learned maps.  All 206 affected map groups were consistent, and the
update made all 17,159 previously unsupported queries supportable.  It added
one instruction form, 100 affine maps, and 1,504 basis rows, bringing the
repository to 369 forms, 3,528 maps, and 56,461 basis rows.}

\revisioncolor{Processing the 24 CUBINs in three different orders produced
the same repository, and replaying them made no further changes.  After
serialization and reload, the updated encoder passed strict round-trip on all
24 CUBINs used for the update.  These results empirically demonstrate that the
continual-training procedure is order independent across the tested batch
orders and idempotent under replay.}

\section{Related Work}

Earlier CUDA assemblers recover encodings at narrower scopes.
MaxAs and TuringAs implement manually recovered encoding
rules~\cite{maxas,DBLP:conf/ppopp/YanWC20}.  Decoding CUDA Binary automates
instruction-format recovery~\cite{DBLP:conf/cgo/HayesHHCZ19}, while CuAssembler
learns arithmetic encoding relations from disassembly--binary
pairs~\cite{cuassembler}.  CuAsmRL uses such an assembler inside a
SASS-scheduling loop~\cite{DBLP:conf/cgo/HeY25}.  \tool{} differs by learning
vector-valued affine maps over $\gf$, using exact Gaussian elimination to
compact observations and test row-span support, and using a shared pipeline
with target-specific profiles for Hopper SM90/SM90a, Blackwell SM100, and
Rubin SM107.

\tool{} empirically establishes that the 5 Blackwell SM targets
share the same SASS instruction-encoding scheme.  Jia et al. found analogous
continuity from Volta to Turing: Turing retained Volta's encodings for common
instructions~\cite{jia-turing-t4-2019}.  Together, these independent findings
suggest that NVIDIA has repeatedly preserved instruction encodings across
related targets and generations.

\tool{} encodes explicitly supplied scheduler-control fields but
does not yet synthesize or optimize them.  Automatic scheduling is a
sequence-level problem, separate from its instruction-local $\gf$ encoder.  It
requires target-specific latency models, dependency analysis,
scoreboard-barrier allocation, and hardware validation.  MaxAs provided
rule-based scheduling for Maxwell, while CuAssembler leaves control fields to
the programmer~\cite{maxas,cuassembler}.  CuAsmRL searches locally reordered
schedules while preserving the original control fields and rejecting swaps
that violate modeled register, barrier, latency, or synchronization
constraints~\cite{DBLP:conf/cgo/HeY25}.
Extending \tool{} with a target-aware scheduler remains future work.

\section{Conclusion}

\tool{} learns exact 128-bit NVIDIA \sass{} encoders as affine
maps over $\gf$.  Streaming Gaussian elimination builds compact bases, detects
inconsistent evidence, and rejects unsupported inputs.
To our knowledge, \tool{} is the first open-source \sass{}
assembler to support instruction encoding and \cubin{} reconstruction for
Rubin SM107.
One jointly trained Rubin encoder also supports SM107, SM107a, and SM107f.
Joint training across 5 Blackwell SM targets provides strong
empirical evidence that NVIDIA uses a common SASS encoding scheme for
instructions shared among these targets.
\newrevisioncolor{For Rubin, continual training made all 17,159 previously
unsupported cuTile and GROMACS queries encodable, adding 1,504 linearly
independent basis rows---the minimum required by the affected map groups.}
\tool{} provides an open-source assembly backend for \sass{} coding agents,
hand-tuned kernels, and microbenchmarking recent NVIDIA GPU architectures.

\bibliography{hopper_assembler2_technical_report}
\bibliographystyle{mlsys2025}

\appendix
\section{Supported Targets}
\label{app:supported-targets}

Table~\ref{tab:target-support} lists the evaluated targets.
Supported targets have a trained encoder and completed byte-exact round-trip
evaluation at the coverage reported in Section 4.  Opcodes count distinct supported mnemonics,
while forms count distinct mnemonic-and-ordered-operand-type structures.
SM90/SM90a share one encoder.
SM100, SM100f, SM100a, SM103, and SM103a share one Blackwell encoder.
SM107, SM107f, and SM107a share one Rubin encoder.

\begin{table}[ht]
\centering
\caption{Exact target support in the evaluated implementation.}
\label{tab:target-support}
\scriptsize
\setlength{\tabcolsep}{2pt}
\begin{tabular}{@{}lllrr@{}}
\toprule
Target(s) & Architecture & Status & Opcodes & Forms \\
\midrule
\texttt{sm\_90/90a} & Hopper    & Supported   & 140 & 389 \\
\shortstack[l]{\texttt{sm\_100/100f/100a}\\\texttt{sm\_103/103a}}
                      & Blackwell & Supported   & 153 & 388 \\
\texttt{sm\_107/107f/107a} & Rubin & Supported & 153 & 369 \\
\bottomrule
\end{tabular}
\end{table}

Extending \tool{} to additional SM architectures remains future work.  Its
learning and training pipeline is shared across targets.
Planned future targets include SM110*, SM120*, and SM121*.

\section{Streaming Gaussian Elimination Example}
\label{app:streaming-elimination-example}

\begingroup
\setlength{\jot}{1pt}
\setlength{\parskip}{4pt}
\setlength{\abovedisplayskip}{2pt}
\setlength{\belowdisplayskip}{2pt}
\setlength{\abovedisplayshortskip}{0pt}
\setlength{\belowdisplayshortskip}{1pt}
This simplified example uses four-bit feature vectors and
8-bit encoding words to show each elimination step.  \tool{} applies the same
algorithm to larger feature vectors and 128-bit instruction words.  Let
$q=[q_3,q_2,q_1,q_0]$ be the input feature vector and $q_{\mathrm{res}}$ its
current residual, initially equal to $q$.  For a nonzero residual, the pivot
$k=\operatorname{msb}(q_{\mathrm{res}})$ is the index of its highest set bit,
with $0\le k\le3$.  The basis table is indexed by pivot position $k$.
Each stored entry $B[k]=(r_k,z_k)$ contains a basis vector $r_k$ with pivot
$k$ and its corresponding reduced output word $z_k$.

The four observations arrive in this order:
\begin{center}
\small
\begin{tabular}{@{}ccl@{}}
\toprule
Observation & $q$ & $y$ \\
\midrule
1 & \texttt{1100} & \texttt{10011111} \\
2 & \texttt{1000} & \texttt{10100101} \\
3 & \texttt{1010} & \texttt{01100110} \\
4 & \texttt{0110} & \texttt{11111001} \\
\bottomrule
\end{tabular}
\end{center}

Observation 1 has highest set bit $q_3=1$, so its pivot is
$k=3$.  Because the basis has no row for this pivot, training stores
\begin{equation}
B[3]=(\texttt{1100},\texttt{10011111}).
\label{eq:toy-first-pivot}
\end{equation}
Observation 2 initially also has $k=3$.  XORing the stored pair updates both
residuals and exposes a new pivot:
\begin{equation}
\begin{aligned}
&(\texttt{1000},\texttt{10100101})\oplus B[3]\\
&\qquad=(\texttt{0100},\texttt{00111010}),\quad k:3\rightarrow2.
\end{aligned}
\label{eq:toy-second-reduction}
\end{equation}
Before storing the new pair at pivot 2, training XORs it into
$B[3]$ to clear bit 2 from that existing row:
\begin{equation}
\mbox{\fontsize{8}{10}\selectfont$\displaystyle
\begin{aligned}
B[3]&\leftarrow(\texttt{1100},\texttt{10011111})
    \oplus(\texttt{0100},\texttt{00111010})\\
    &=(\texttt{1000},\texttt{10100101}).
\end{aligned}
$}
\label{eq:toy-backward-clear}
\end{equation}
Training then stores the new pair as
$B[2]=(\texttt{0100},\texttt{00111010})$.  The basis now has rank 2, with
entries $B[3]$ and $B[2]$.

Observation 3 follows pivot 3 and then exposes pivot 1:
\begin{equation}
\begin{aligned}
&(\texttt{1010},\texttt{01100110})\oplus B[3]\\
&\qquad=(\texttt{0010},\texttt{11000011}),\quad k:3\rightarrow1.
\end{aligned}
\label{eq:toy-third-reduction}
\end{equation}
Training stores this pair as $B[1]$.  The resulting rank-3 basis is
\begin{equation}
\begin{aligned}
B[3]&=(\texttt{1000},\texttt{10100101}),\\
B[2]&=(\texttt{0100},\texttt{00111010}),\\
B[1]&=(\texttt{0010},\texttt{11000011}).
\end{aligned}
\label{eq:toy-final-basis}
\end{equation}

Observation 4 visits pivots $2\rightarrow1$:
\begin{equation}
\begin{aligned}
(\texttt{0110},\texttt{11111001})\oplus B[2]
  &=(\texttt{0010},\texttt{11000011}),\\
\phantom{(\texttt{0110},\texttt{11111001})}\oplus B[1]
  &=(\texttt{0000},\texttt{00000000}).
\end{aligned}
\label{eq:toy-redundant-reduction}
\end{equation}
Both residuals are zero, so this observation is redundant and is discarded.
If its output had instead been \texttt{11111000}, the final residual would be
$(\texttt{0000},\texttt{00000001})$: identical features would imply two
different encodings, so training would report a contradiction.

Training therefore retains three basis pairs, $B[3]$, $B[2]$,
and $B[1]$, forming a rank-3 model.  The fourth observation adds no new
information.  We now use this fixed basis to encode unseen inputs.

At encoding time, \tool{} receives only a query feature vector and
initializes an output accumulator to zero.  The unseen query \texttt{1110} does
not appear among the four training observations.  It visits pivots
$3\rightarrow2\rightarrow1$:
\begin{equation}
\begin{array}{c|c|c}
\text{Operation} & q_{\mathrm{res}} & y_{\mathrm{acc}}\\ \hline
\text{initial}   & \texttt{1110} & \texttt{00000000}\\
\oplus B[3]      & \texttt{0110} & \texttt{10100101}\\
\oplus B[2]      & \texttt{0010} & \texttt{10011111}\\
\oplus B[1]      & \texttt{0000} & \texttt{01011100}
\end{array}
\label{eq:toy-inference-trace}
\end{equation}
The zero feature residual proves that this unseen query is
supported; the final accumulator value,
$y_{\mathrm{acc}}=\texttt{01011100}$, is the query's 8-bit encoding.

For the unseen query $q=\texttt{1001}$, reduction with
$r_3=\texttt{1000}$ leaves
\begin{equation}
q_{\mathrm{res}}
  =\texttt{1001}\oplus\texttt{1000}=\texttt{0001}.
\label{eq:toy-unsupported-inference}
\end{equation}
The residual has pivot $0$, but the trained basis in
Eq.~\eqref{eq:toy-final-basis} contains only rows with pivots $3$, $2$, and
$1$.  Therefore, the residual cannot be eliminated.  \tool{} rejects the
query, discards the partial accumulator, and emits no encoding.
\par\endgroup

\section{End-to-End Instruction Lifecycle}
\label{app:instruction-lifecycle}

\begin{table*}[t]
\centering
\caption{Five of the encoding map's 16 stored basis pairs used
for this HMMA example.  Here $v_i[b]$ denotes bit $b$ of parsed value $v_i$;
each $z_k$ is shown as a 128-bit hexadecimal word.}
\label{tab:hmma-basis}
\scriptsize
\begin{tabular*}{\textwidth}{@{\extracolsep{\fill}}rll@{}}
\toprule
$k$ & Active features in $r_k$ &
  $z_k$ (128-bit hexadecimal) \\
\midrule
259 & $v_1[2]\oplus v_2[2]\oplus v_4[2]$ &
  \texttt{00000000000000040000000004040000} \\
\addlinespace[1.5pt]
199 & $v_3[6]$ &
  \texttt{00000000000000000000004000000000} \\
\addlinespace[1.5pt]
196 & $v_3[3]$ &
  \texttt{00000000000000000000000800000000} \\
\addlinespace[1.5pt]
135 & $v_2[6]$ &
  \texttt{00000000000000000000000040000000} \\
\addlinespace[1.5pt]
3 & $1\oplus v_0[0]\oplus v_0[1]\oplus v_0[2]$ &
  \texttt{0400000000041800000000000000723c} \\
\bottomrule
\end{tabular*}
\end{table*}

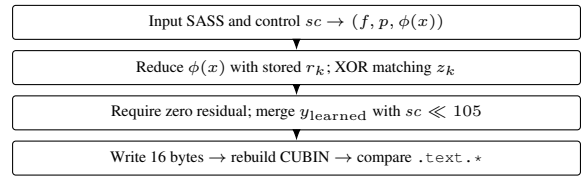
\begin{figure}[ht]
  \centering
  \begin{tikzpicture}[
      node distance=1.35mm,
      life/.style={draw,
                   rounded corners=1.2pt, align=center,
                   minimum width=0.91\columnwidth, minimum height=4.5mm,
                   inner sep=2.2pt, font=\tiny},
      phase/.style={font=\scriptsize\bfseries,
                    anchor=west},
      lifearrow/.style={-{Latex[length=1.5mm]},
                       line width=0.45pt}]
    \node[phase] (trainlabel) at (-0.455\columnwidth,0)
      {Offline training};
    \node[life, below=0.7mm of trainlabel.south west, anchor=north west]
      (source) {CUBIN $\rightarrow$ extract $(x,y_{\mathrm{learned}})$};
    \node[life, below=of source] (feature) {Parse and vectorize
      $\rightarrow (f,p,\phi(x))$};
    \node[life, below=of feature] (train) {Gaussian elimination
      $\rightarrow$ reduced basis $\{(r_k,z_k)\}$};
    \node[phase, below=2.0mm of train.south west, anchor=north west]
      (runlabel) {Runtime assembly};
    \node[life, below=0.7mm of runlabel.south west, anchor=north west]
      (input) {Input SASS and control $\mathit{sc}$
      $\rightarrow (f,p,\phi(x))$};
    \node[life, below=of input] (encode) {Reduce $\phi(x)$ with stored $r_k$;
      XOR matching $z_k$};
    \node[life, below=of encode] (merge) {Require zero residual; merge
      $y_{\mathrm{learned}}$ with $\mathit{sc}\ll105$};
    \node[life, below=of merge] (verify) {Write 16 bytes $\rightarrow$ rebuild
      CUBIN $\rightarrow$ compare \texttt{.text.*}};
    \draw[lifearrow] (source) -- (feature);
    \draw[lifearrow] (feature) -- (train);
    \draw[lifearrow] (input) -- (encode);
    \draw[lifearrow] (encode) -- (merge);
    \draw[lifearrow] (merge) -- (verify);
  \end{tikzpicture}
  \caption{Two phases in the instruction lifecycle.  Training
  constructs a reusable exact basis.  Runtime assembly queries it and fails
  if the feature vector is unsupported.}
  \label{fig:instruction-lifecycle}
\end{figure}

This appendix illustrates how \tool{} learns and applies an
exact instruction encoder.  It separates offline solution of the encoding
equations from runtime assembly, then follows one SM100 Tensor
Core instruction from SASS text to its final 16-byte word.  The example makes
the model in Section~2 concrete by showing parsed features, stored
$\mathbb{F}_2$ coefficients, learned-output reconstruction, scheduler-control merge,
and CUBIN verification.  Figure~\ref{fig:instruction-lifecycle} summarizes the
two phases.  The observation comes from NVIDIA cuDNN 9.25.0.15, CUBIN SHA-256
prefix \texttt{02618e0d}, at program counter \texttt{0x54d0}.  The runtime
input is
\par\vspace{2pt}
{\centering\small
\texttt{[B--23--:R-:W-:-:S01]}\\[-1pt]
\texttt{HMMA.16816.F32.BF16 R4, R68.reuse, R72, R4;}\par}
\vspace{2pt}
\noindent HMMA has four register operands in the order destination $D$,
source $A$, source $B$, and accumulator $C$, and computes
$D\leftarrow A B+C$.  In this example they are respectively
\texttt{R4}, \texttt{R68}, \texttt{R72}, and \texttt{R4}.
The \texttt{.reuse} suffix on \texttt{R68} marks the first source
operand, $A$, for operand reuse, allowing its value to be retained for
subsequent instructions~\cite{maxas-sgemm}.  The reuse flag is
encoded according to the source-operand position, not the register number.

The bracketed prefix supplies scheduler control
$\mathit{sc}=\texttt{0x067f1}$.
\tool{}'s SASS parser produces form
$f=\texttt{HMMA\_R\_R\_R\_R}$, modifier tuple
$m=(\texttt{HMMA},\texttt{16816},\texttt{F32},\texttt{BF16},
\texttt{first-source reuse})$, and values $v=(7,4,68,72,4)$.
Their operand roles are
\begin{equation}
  \mbox{\fontsize{8}{10}\selectfont$\displaystyle(v_0,v_1,v_2,v_3,v_4)
  =(\texttt{PT},D,A,B,C)
  =(7,4,68,72,4).$}
  \label{eq:hmma-parsed-values}
\end{equation}
Equation
\ref{eq:hmma-feature-vector} gives the resulting
$\phi(x)\in\mathbb{F}_2^{321}$.

Equation~\ref{eq:hmma-feature-vector} supplies the input side
of each training equation, while the observed instruction word supplies $y$
in Equation~\ref{eq:encoding-model}.  Gaussian elimination applies identical
XOR row operations to both sides and produces stored pairs $(r_k,z_k)$.  Here
$r_k\in\gf^d$ is a reduced feature vector whose highest active feature $k$ is
its pivot, and $z_k\in\gf^{128}$ is the correspondingly reduced instruction
word.  Thus every pair satisfies
\begin{equation}
  \underbrace{r_k}_{1\times d}
  \underbrace{W}_{d\times128}
  =\underbrace{z_k}_{1\times128}.
  \label{eq:basis-pair-invariant}
\end{equation}
To encode an input SASS instruction $x$, \tool{} reduces
$\phi(x)$ using the stored $r_k$ rows and XORs the corresponding $z_k$
words into the output.

For the learned encoding map used by this HMMA example,
16 observations produce a rank-16 basis.
\tool{} stores this learned map as reduced basis pairs
$(r_k,z_k)$, rather than a dense $321\times128$ coefficient matrix.
Table~\ref{tab:hmma-basis} shows the
five of the shared map's 16 stored basis pairs selected while
reducing this
HMMA query.  The remaining 11 pairs have reduction coefficient zero for this
query.

For this input, the active feature vector decomposes as
\begin{equation}
  \phi(x)
  =r_{259}\oplus r_{199}\oplus r_{196}\oplus r_{135}\oplus r_3.
  \label{eq:hmma-feature-reduction}
\end{equation}
The residual feature vector becomes zero, proving that the
query is supported.  XORing the five coefficients gives the learned
non-scheduler instruction bits:
{\footnotesize
\begin{equation}
\begin{aligned}
y_{\mathrm{learned}}
 &=z_{259}\oplus z_{199}\oplus z_{196}\oplus z_{135}\oplus z_3\\
 &=\texttt{0x0400000000041804000000484404723c}.
\end{aligned}
\label{eq:hmma-learned-reconstruction}
\end{equation}}

Scheduler control is represented separately from the learned instruction bits.  The
SM100 profile places its 17 bits at positions 105--121:
{\footnotesize
\begin{equation}
\begin{aligned}
\mathit{sc}\ll105
 &=\texttt{0x00cfe200000000000000000000000000},\\
y_{\mathrm{full}}
 &=y_{\mathrm{learned}}\oplus(\mathit{sc}\ll105)\\
 &=\texttt{0x04cfe20000041804000000484404723c}.
\end{aligned}
\label{eq:hmma-full-word}
\end{equation}}

\tool{} emits $y_{\mathrm{full}}$ as the following 16 little-endian bytes and
writes them into the executable section:
\begin{center}
\small\ttfamily
3c 72 04 44 48 00 00 00\\
04 18 04 00 00 e2 cf 04
\end{center}
The rebuilt CUBIN then passes the same byte-exact \texttt{.text.*} comparison
used by the whole-corpus round-trip evaluation.

The same $(f,p,v,\phi)$ representation also accommodates
Blackwell's \texttt{UTCOMMA} instructions with descriptor and tensor-memory
operands~\cite{nvidia-binary-utils-sm90}.  The following SM100f
example comes from \texttt{flashinfer-cubin==0.6.13} (CUBIN SHA-256 prefix
\texttt{04b33a6d}):
\par\smallskip
{\centering\footnotesize\ttfamily
\begin{tabular}{@{}l@{}}
UTCOMMA.2CTA.4X gdesc[UR42],\\
\phantom{UTCOMMA.2CTA.4X }gdesc[UR52], tmem[UR73],\\
\phantom{UTCOMMA.2CTA.4X }tmem[UR22], idesc[UR23],\\
\phantom{UTCOMMA.2CTA.4X }tmem[UR26], UPT;
\end{tabular}\par}
\smallskip
\tool{} parses eight values, including the implicit guard,
\begin{equation*}
v=(7,42,52,73,22,23,26,7),
\end{equation*}
yielding a feature vector $\phi(x)\in\mathbb{F}_2^{513}$.
Section~\ref{sec:roundtrip-experiments} reports byte-exact validation across
Blackwell and Rubin instruction corpora.

\section{DEPBAR bits omitted by \texttt{nvdisasm}}
\label{app:depbar-opaque}

\texttt{DEPBAR} is a SASS dependency-barrier
instruction~\cite{nvidia-binary-utils-sm90}.  The example disassembles as
\texttt{DEPBAR.LE SB5, 0x2}.  For this instruction in the
\texttt{DEPBAR\_SB\_II} context, \texttt{nvdisasm} omits bits 122 and 124 from
its text output.  These bits lie in the reuse/control region (bits 122--125),
adjacent to scheduler-control bits 105--121.  One SM100 example makes the
ambiguity concrete.  After clearing scheduler-control bits, two
observations yield
\par\smallskip
{\centering\scriptsize\ttfamily
\begin{tabular}{@{}r@{ = }l@{}}
word A & 0x00000000000000000000d0800000791a\\
word B & 0x14000000000000000000d0800000791a\\
nvdisasm(A) & DEPBAR.LE SB5, 0x2;\\
nvdisasm(B) & DEPBAR.LE SB5, 0x2;\\
A XOR B & 0x14000000000000000000000000000000
\end{tabular}\par}
\smallskip
The XOR has ones only at bits 122 and 124.  Thus \texttt{nvdisasm}'s text
output cannot determine these two bit values.  This is a \texttt{nvdisasm}
output limitation, not a limitation of \tool{}'s encoding model.  We found 6
ambiguous locations on SM100 and 187 on SM103, but none in the collected
SM100a, SM100f, or SM103a corpora.

\tool{} handles this \texttt{nvdisasm} limitation explicitly.
It applies one target-independent opaque mask to the context.  The mask
excludes bits 122 and 124 from the learned instruction bits.  During CUBIN-to-assembly
export, \tool{} records the original high-half state in enriched
\texttt{:H...} state.  Assembly restores the mask-selected bits after learned
encoding.  This preserves byte-exact reconstruction without using the SM
target as a feature or learned-map selector.

Future work will determine whether these bits have
architectural meaning.  If so, \tool{} can expose them as parsed modifiers or
features and encode them like other instruction fields.  Otherwise, explicit
opaque-state preservation remains necessary for byte-exact reconstruction.

\end{document}